\documentclass[11pt,letterpaper]{article}

\usepackage[margin=1in]{geometry}
\usepackage[numbers,sort&compress]{natbib}
\usepackage[utf8]{inputenc}
\usepackage[T1]{fontenc}
\usepackage{lmodern}
\usepackage{url}
\usepackage{booktabs}
\usepackage{graphicx}
\usepackage{amsfonts}
\usepackage{nicefrac}
\usepackage{microtype}
\usepackage{xcolor}
\usepackage{array}
\usepackage{amsmath}
\usepackage{placeins}
\usepackage{fvextra}
\usepackage{titling}
\usepackage{adjustbox}
\usepackage{etoolbox}
\usepackage{hyperref}

\hypersetup{
  colorlinks=true,
  linkcolor=blue!50!black,
  citecolor=blue!50!black,
  urlcolor=blue!50!black
}

\newcommand{\term}[1]{\textsc{#1}}
\newcolumntype{L}[1]{>{\raggedright\arraybackslash}p{#1}}
\newcolumntype{C}[1]{>{\centering\arraybackslash}p{#1}}

\preauthor{\begin{center}\normalsize\begin{tabular}{c}}
\postauthor{\end{tabular}\end{center}}
\date{}

\BeforeBeginEnvironment{tabular}{\begin{adjustbox}{max width=\linewidth}}
\AfterEndEnvironment{tabular}{\end{adjustbox}}

\DefineVerbatimEnvironment{PromptBox}{Verbatim}{
  breaklines=true,
  breakanywhere=true,
  fontsize=\small,
  frame=single,
  framesep=2mm,
  rulecolor=\color{black!35}
}

\title{EchoPath: Execution-Level Replayable Memory for GUI Agents}
\author{
  Yao Zhao\textsuperscript{1} \quad
  Aditya Shanmugham\textsuperscript{3} \quad
  Swastik Roy\textsuperscript{3} \quad
  Yanxun Xu\textsuperscript{1,2} \\
  \textsuperscript{1}Department of Applied Mathematics and Statistics, Johns Hopkins University \\
  \textsuperscript{2}Division of Quantitative Sciences, Department of Oncology, \\
  Johns Hopkins University School of Medicine \\
  \textsuperscript{3}Amazon AGI
}

\begin{document}
\maketitle

\begin{abstract}
Computer-use agents increasingly operate browsers, software, and desktop applications via CLI or API portals, but graphical user interface (GUI) still plays an important role in common industrial production scenarios. GUI agents commonly employ fresh observe-plan-ground-act loops, which is inefficient for enterprise tasks that repeatedly update records, process forms, configure tools, and export reports. We introduce EchoPath, a model-agnostic harness that converts artifact-validated GUI trajectories into standardized, parameter-controlled callable memories, analogous to Model Context Protocol (MCP)-style tool calls rather than unstructured experience records. Each memory stores task-intent keys, application and state preconditions, flexible input parameters, GUI evidence, validation provenance, and lifecycle state, so the host agent invokes a targeted procedure only when it can be deterministically replayed in the current runtime. The core mechanism enabling replay is an image-based target-reaiming algorithm that treats stored coordinates as visual evidence, matches the remembered GUI target against the current screen, and emits corrected operation coordinates before execution. During replay, EchoPath rebinds only declared modifiable inputs and rejects ambiguous or incompatible steps to bounded grounding repair or fresh planning. In experiments with real computer-use tasks, EchoPath reduced median token cost by more than 90\% and median execution time by about 60\%. These results support a bounded form of enterprise GUI memory: validated execution experience can become a controllable callable asset for recurrent work rather than only context for another reasoning pass.
\end{abstract}

\section{Introduction}
Although command-line interfaces (CLIs) and application programming interfaces (APIs) provide efficient control for many computing tasks, many real-world workflows cannot be completed through terminal commands alone. Agents therefore also need to operate graphical user interfaces (GUIs) across browsers, productivity suites and desktop applications. This requirement has motivated operating-system benchmarks, including OSWorld, Windows Agent Arena and AndroidWorld, that evaluate agents by external task outcomes in realistic computing environments \cite{xie2024osworld,bonatti2025windowsarena,rawles2025androidworld,abhyankar2025osworldhuman}. Current computer-use agents usually solve each GUI task through an observe-plan-ground-act loop. This loop is necessary for novel tasks, but it is inefficient when the agent has already completed an equivalent procedure and conducts similar tasks repeatedly. In enterprise settings, computer-use agents are often assigned recurrent tasks, such as updating records, processing forms, reconciling spreadsheets, moving information across tools, and exporting reports. In these production scenarios, cost, latency, auditability, privacy and operational control matter as much as single-task success.

To improve the efficiency on recurrent tasks, existing memory systems often retrieve prior trajectories, reflections, workflow summaries, or skills as context for another model-driven plan \citep{shinn2023reflexion,zhao2024expel,kagaya2024rap,wang2024awm,fang2025memp,han2026legomem}, but it is not directly executable or easily auditable. Classical record-and-replay systems execute prior actions more directly, but raw coordinates and brittle scripts fail under window movement, resolution changes, layout drift, application updates, or repeated interface elements \citep{yeh2009sikuli,barman2016ringer,dong2022webrobot}. Modern agentic record-and-replay workflows~\citep{openai2026recordreplay} can bind actions to interface state rather than screen coordinates alone, but this flexibility is typically constrained by the GUI elements exposed through the operating system's accessibility tree. Considering the landscape of GUI agent (Supplementary Material, Section~A), GUI agent memory for recurrent tasks therefore needs a representation that is more executable than prompt memory and more state-aware, portable and inspectable than coordinate or accessibility-tree replay.

We introduce EchoPath, a model-agnostic and environment-agnostic schema and harness for execution-level replayable memory. The central idea is to convert a validated GUI trajectory into a standardized, parameter-controlled callable memory, analogous to a Model Context Protocol (MCP)-style tool call rather than an unstructured experience record. Each memory exposes a bounded callable workflow: 1) task-intent keys select the procedure; 2) application and state preconditions gate whether it can run; 3) typed action parameters identify which inputs may be modified; 4) visual evidence binds pointer operations to the current screen; and 5) validation and lifecycle metadata govern promotion, repair, branching, deprecation, and quarantine. The host agent therefore invokes a targeted memory only when the stored procedure is compatible with the current task and workspace.

EchoPath executes a selected memory deterministically while still allowing controlled adaptation. Flexible text-like inputs can be rebound only at declared parameter paths, while fixed navigation, save actions, hotkeys, waits, and structural operations remain unchanged. Pointer actions are not copied as absolute coordinates. Instead, stored target crops and coordinate context are matched against the current screenshot to re-aim the operation before execution via an imaged-based target-reaiming (IBTR) algorithm. If a target is ambiguous, a state precondition fails, or a non-flexible field would need to change, EchoPath rejects direct replay and falls back to bounded grounding repair or ordinary agentic planning for robustness. We evaluate EchoPath with a paired two-pass study on OSWorld-Verified tasks. Against a trajectory-prompt memory baseline, EchoPath maintains comparable task-completion reliability while substantially reducing repeated model use and execution time. Component studies further examine memory retrieval under repository clutter, visual re-aiming under resolution changes, replay start-state gating, flexible-input rebinding, and storage footprint.

Our contributions are fourfold. First, we formulated GUI-level memory as a standardized callable schema for deterministic replay with controlled parameters. Second, we proposed IBTR algorithm that bridges the gap between GUI action and replayable memory bypassing grounding reasoning step. Third, we introduced a model-agnostic harness that combines retrieval, compatibility gating, flexible-input rebinding, replay-time GUI re-aiming, deterministic execution, and bounded fallback through a common action boundary. Fourth, we defined a governed lifecycle for executable memories, including artifact-gated promotion, branching, repair, merge, deprecation, quarantine, and lineage tracking. The OpenPath package is provided at: \url{https://github.com/JackZhao1998/EchoPath.git}

\section{Methods}

\begin{figure}[ht]
\centering
\includegraphics[width=\linewidth]{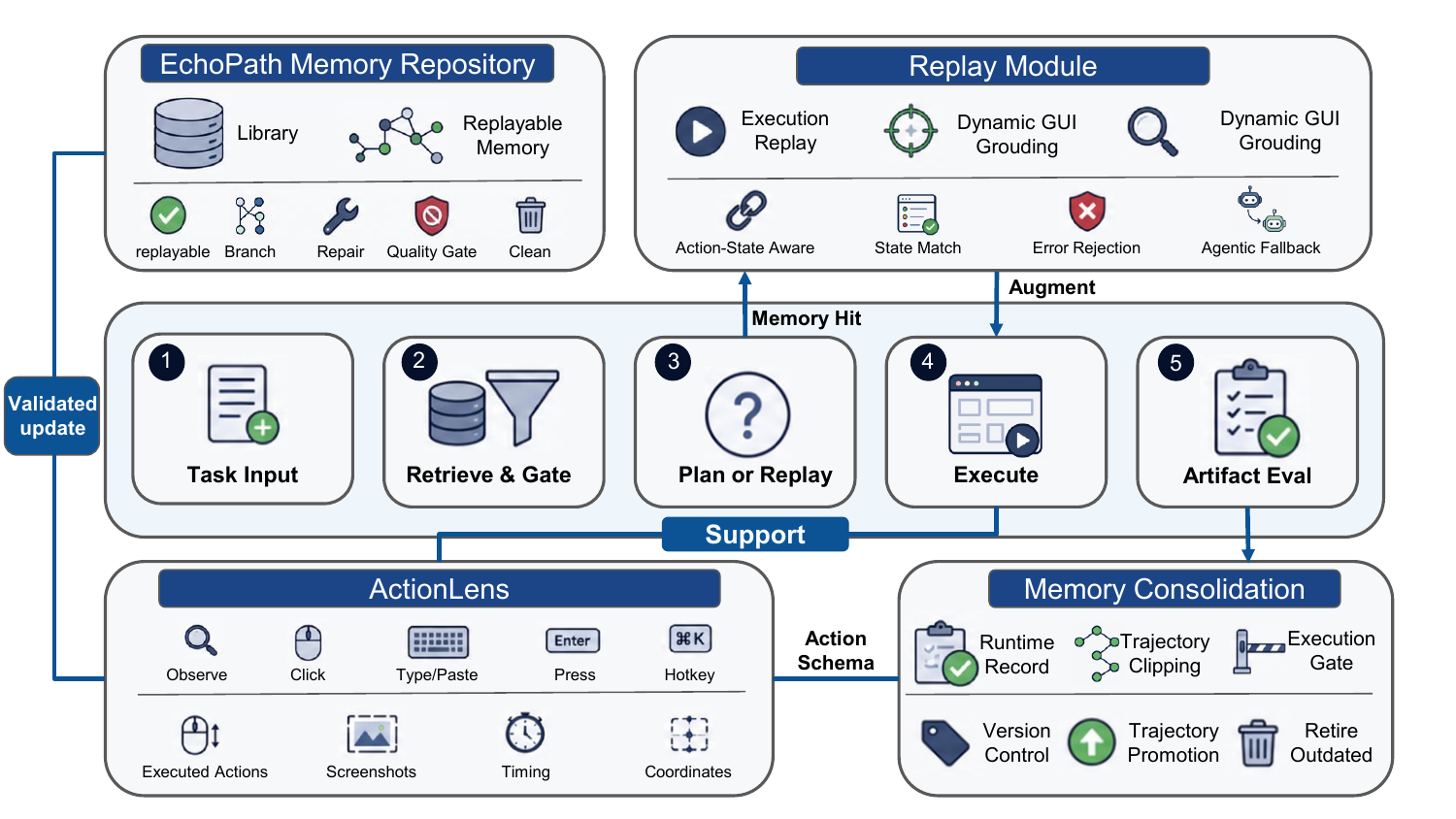}
\caption{EchoPath architecture overview.}
\label{fig:echopath-architecture}
\end{figure}

EchoPath formulates recurrent GUI operation as a five-stage, memory-based control loop: task input, retrieval and gating, plan-or-replay selection, execution, and artifact evaluation (Figure~\ref{fig:echopath-architecture}). It includes a dedicated GUI operation wrapper ActionLens supporting both freshly planned execution following a ReAct style and replay through a common action schema. It enables agent to take actions including observation, clicking, text entry, key presses, hotkeys, scrolling, and waiting, and records the executed actions, screenshots, target GUI crops, timing, and coordinate context needed for diagnosis and artifact evaluation. Technical details of ActionLens are reported in Supplementary Materials Section B.1.    

For a given task input, the system queries the EchoPath Memory Repository for a replayable memory whose task intent, application context, state assumptions, validation evidence, and action schema are compatible with the current request. Then, plan-or-replay stage determines how the task reaches execution. A compatible memory hit invokes EchoPath action replay, which combines execution replay with action-state awareness, state matching, replay-time GUI grounding, error rejection, and a tiered fallback policy. Based on the matched memory, replay module materializes the stored action path as current GUI actions, while the image-based target-reaiming (IBTR) algorithm uses stored visual and coordinate evidence to re-aim pointer actions when the current interface satisfies the episode's preconditions therefore bypasses traditional GUI grounding steps. If a step cannot be bound locally, EchoPath may first invoke a bounded agent-grounding repair for that memorized step. If the episode is structurally incompatible, it rejects direct replay and returns control to memory-guided or ordinary agentic planning. The replay path thus bypasses fresh per-step planning and grounding when the stored episode remains compatible, while the planning fallback preserves flexibility and supports robust task completion when direct replay is no longer appropriate. For a completed task, the artifact evaluator determines whether the task outcome is valid. Memory Consolidation then combines the runtime record with trajectory clipping, execution and quality gates, version control, trajectory promotion, and retirement of outdated paths. In addition, the memory consolidation module also classifies actions as "fixed" type such as clicking, pressing hotkeys or "flexible" type where dynamic input could be inserted such as typing into search box or pasting targeted values. This action classification labeling enables EchoPath to handle tasks with same action trajectories with different contextual information including updating client information or organizing receipts from different sources. Only a validated update is written back to the repository, where replayable memories can be branched, repaired, quality-gated, or cleaned. Memory retrieval and consolidation modules are reported in detail at Supplementary Materials Section B.2.

\subsection{Problem formulation}
We formulate EchoPath around three objects: a GUI operation task, a replayable memory, and a replay task. A GUI operation task is denoted by $\tau=(q,\xi,V)$, where $q$ is the user instruction, $\xi$ describes the desktop context such as application, workspace, file state, and allowed budget, and $V$ is an external artifact evaluator. At step $t$, the desktop has latent state $x_t$ and observation $o_t=O(x_t)$. A GUI action is written as $a_t=(\alpha_t,\theta_t)$, where $\alpha_t$ is an action primitive and $\theta_t$ contains normalized parameters such as text, key names, scroll distance, coordinates, or wait duration. The ActionLens action space is
\[
\mathcal{A}=\{\mathrm{observe},\mathrm{click},\mathrm{move},\mathrm{type},\mathrm{paste},
\mathrm{press},\mathrm{hotkey},\mathrm{scroll},\mathrm{wait}\}.
\]
Executing $a_t$ through ActionLens produces a new state and an evidence record:
\[
x_{t+1},z_t = W(x_t,a_t),
\]
where $W$ is the wrapped GUI execution boundary and $z_t$ stores before/after observations, screenshots, timing, executor output, error status, and coordinate context. A completed first run is therefore a trace
\[
\Gamma = \left(\tau, x_0, (o_t,a_t,z_t)_{t=1}^{T}, y_T\right),
\]
where $y_T$ is the final artifact or task output. The run is eligible for memory construction only when the evaluator passes, $V(y_T,\tau)=1$.

A replayable memory is an execution-level episode derived from an evaluated trace. We write an episode as
\[
m=(K_m,u_m,\Pi_m,A_m,\mathcal{E}_m,B_m,v_m,r_m,\ell_m).
\]
Here $K_m$ is the set of retrieval keys, $u_m$ is the application or environment label, $\Pi_m$ stores state preconditions and expected effects, $A_m=(a_1,\ldots,a_T)$ is the executable action program, $\mathcal{E}_m=(e_1,\ldots,e_T)$ stores visual and state evidence for the actions, $B_m$ is an optional set of flexible action-parameter bindings, $v_m$ is artifact-validation evidence, $r_m$ is a reasoning or consolidation report, and $\ell_m$ is the lifecycle state. Only episodes whose lifecycle state is active are exposed for default replay retrieval.

A replay task asks whether a new task $\tau'$ can reuse an active episode rather than start fresh GUI control. EchoPath first retrieves and gates a candidate set $\mathcal{C}(\tau')\subseteq\mathcal{M}_{\mathrm{active}}$, where $\mathcal{M}_{\mathrm{active}}$ is the active subset of the memory repository. If a memory $m^\star$ is selected, action replay observes the current replay state $o'_t=O(x'_t)$ and maps each stored action to a current action,
\[
\tilde{a}_t=\rho(a_t,o'_t,e_t,\Pi_{m^\star},B_{m^\star},\xi'),
\]
and executes $\tilde{a}_t$ through the same ActionLens boundary $W$. If retrieval or binding fails, the replay task is rejected and control returns to regular GUI operation. This definition separates three decisions that are often conflated: whether a prior run should become memory, whether a memory should be retrieved for a new task, and whether the retrieved action program can be safely bound to the current GUI state.

\subsection{EchoPath Action Replay}
\label{sec:echoaim}

\begin{figure}[ht]
\centering
\includegraphics[width=\linewidth]{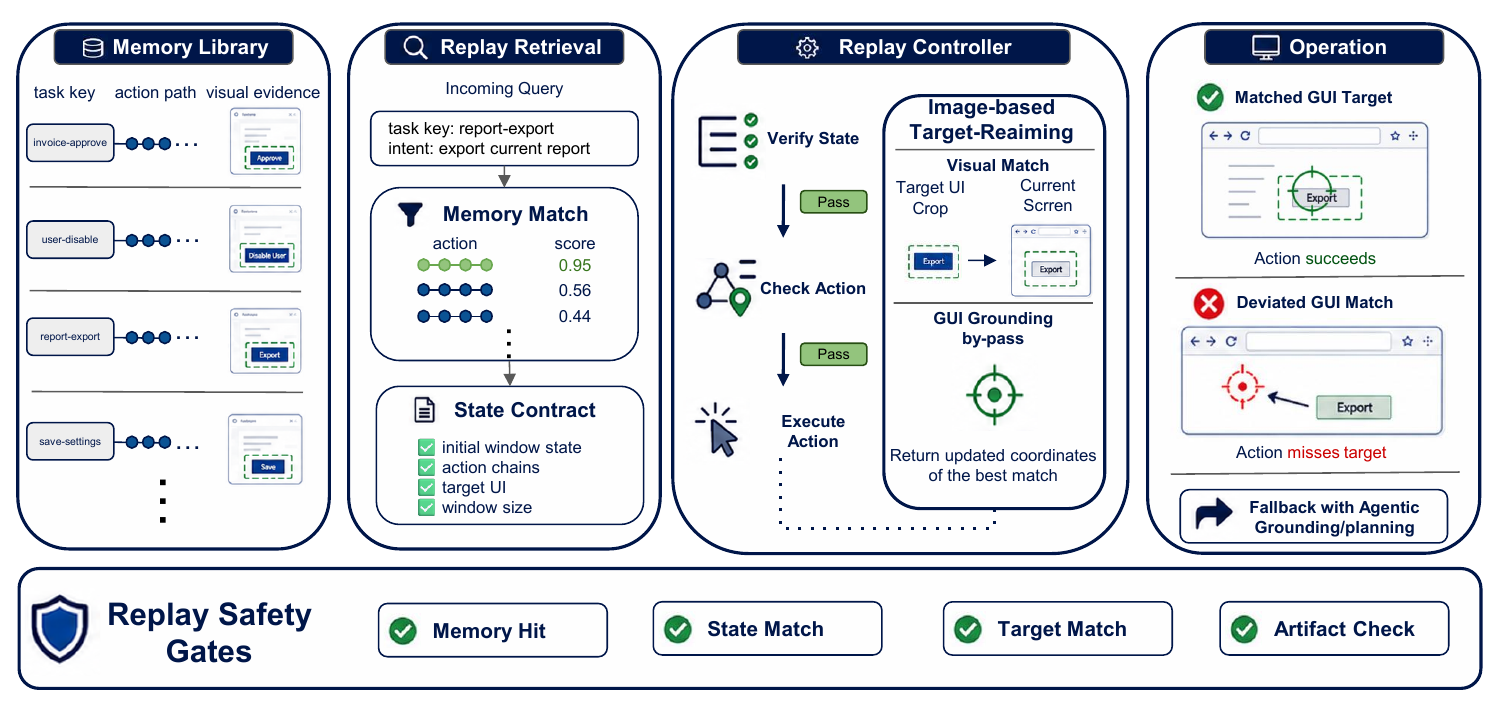}
\caption{EchoPath action replay with automatic GUI target re-aiming and agentic fallback.}
\label{fig:echoaim-reaiming}
\end{figure}

EchoPath action replay replaces fresh per-step GUI planning and grounding with compatibility-gated instantiation of a stored action trajectory. As shown in Figure~\ref{fig:echoaim-reaiming}, its input is a selected memory $m^\star$, a new task $\tau'$, the current desktop state $x_0$, and the retrieval gate report. Its output is either a bound replay program $\widetilde{A}=(\tilde{a}_1,\ldots,\tilde{a}_T)$ or a rejection decision with a cause, such as state mismatch, missing target, parameter conflict, unsupported action, or unsafe action. This procedure is needed because a stored episode is not merely a coordinate script: it is a validated action path whose targets, parameters, visual evidence, and preconditions must be checked against the current GUI before replay.

For actions that require precise coordinates, such as clicks, action replay invokes the image-matching-based target-reaiming algorithm. Its role is to answer a narrower question than ordinary GUI grounding: given a GUI target that was already used successfully in a source action memory, can that same target be located and safely acted on in the current observed state. The stored memory supplies the saved action, the source coordinate frame, the target crop, screen and window metadata, and state preconditions. EchoPath then performs visual matching against the current screenshot, checks whether the current GUI state satisfies the stored preconditions, and binds the matched target to the current coordinate frame. Since naive reuse of the old or linearly scaled coordinate can miss after resizing, window dragging, or layout drift, EchoPath instead treats the coordinate as evidence about a visual target, not as the portable target itself.

We formulate the EchoPath replay module as follows. In a fresh GUI-agent loop, step $t$ typically performs observation, planning, grounding, and execution:
\[
o_t=O(x_t), \qquad
p_t=F_\theta(q,h_{t-1},o_t), \qquad
\hat{a}_t=G_\phi(p_t,o_t), \qquad
x_{t+1},z_t=W(x_t,\hat{a}_t).
\]
Here $F_\theta$ denotes the host model's planning function, $G_\phi$ denotes a GUI-grounding function that binds a proposed step to coordinates or controls, and $h_{t-1}$ is the run history. This loop is necessary for novel tasks, but it repeats expensive reasoning and grounding even when a validated procedure already exists. EchoPath action replay narrows the problem: it does not ask what action should be taken next; it asks whether the already validated action $a_t$ can be instantiated safely in the current observation.

For each stored action, action replay applies a binding operator
\[
\rho(a_t,o'_t,e_t,\Pi_{m^\star},B_{m^\star},\xi')
  \in \{(\tilde{a}_t,\kappa_t,\delta_t),\mathrm{reject}\},
\]
where $\kappa_t$ is a binding confidence and $\delta_t$ records whether the step was copied, parameter-rebound, or visually re-aimed. The evidence item $e_t\in\mathcal{E}_{m^\star}$ includes a component signature, source coordinate frame, target crop, source screenshot, screen size, window metadata, and source-step provenance. When the operator returns a bound action, action replay executes it through the same ActionLens boundary:
\[
x'_{t+1},z'_t=W(x'_t,\tilde{a}_t), \qquad t=1,\ldots,T .
\]
These actions are therefore executed deterministically with coordinates supplied by the target-reaiming algorithm described in Supplementary Material, Section B.3.

If an action were rejected, EchoPath employs either a step-local grounding fallback or task-level planning fallback. This separation is important because a failed replay bind can mean two different things. A local grounding failure means that the remembered step is still plausible, but its immediate target cannot be bound by deterministic coordinate scaling, flexible parameter rebinding, or image matching. Examples include a shifted widget, an ambiguous GUI crop, a stale pointer coordinate, or transient focus drift while the task, application, and prior replay state still satisfy the action's preconditions. In this case, EchoPath can invoke the GUI agent as a bounded grounding repair operator rather than as a new planner:
\[
\hat{a}^{\mathrm{gnd}}_t =
\psi_{\mathrm{ground}}(a_t,e_t,o'_t,\Pi_{m^\star},\xi').
\]
The agent grounding operator $\psi_{\mathrm{ground}}$ receives the memorized action, its visual and state evidence, the current observation, the episode preconditions, and the current workspace context.

A structural failure means that the remembered episode is no longer a safe procedure for the current state, for example because the wrong application is active, a required document state is absent, a non-flexible parameter would need to change, the action primitive is unsupported, or the observed postcondition contradicts the stored transition. The planner then receives the current task and observation and constructs a fresh policy,
\[
\pi_{\mathrm{plan}} =
F_\theta(q',h'_t,o'_t,\chi_{m^\star}),
\]
where $\chi_{m^\star}$ is either empty for ordinary fallback or with prior selected memory as reference for a planner that is piloting from the live screen.

The replay operator above is implemented by a set of core supporting components. ActionLens provides the controller-compatible execution boundary, exposes normalized GUI primitives, and records before/after observations, screenshots, timing, error status, and coordinate context. The memory repository retrieves and gates active episodes using task intent, application, validation, lifecycle, reasoning, and action-schema checks, then consolidates validated traces through promotion, branching, merging, deprecation, or quarantine. Replay-time binding adds image-match target-reaiming, flexible-input reasoning, and bounded fallback: stored coordinates are treated as visual evidence, matched against the current screenshot, rejected when ambiguous, and converted into the current controller frame.

\section{Experiments}

\subsection{Main Experiment Setup}
The main experiment validates whether EchoPath can efficiently conduct GUI operation based pre-consolidated memories. We therefore use a paired two-pass design on OSWorld-Verified tasks. The first pass constructs candidate memories from fresh GUI execution, while the second pass reruns the same task entries and tests whether the expected memory can be retrieved, gated, rebound, and executed. To avoid "coordinate copying" style memory which replays every action simply using the same target coordinately, we intentionally shifted second-pass resolution from 1920$\times$1080 in the first-pass to 1600$\times$900. This resolution shift is deliberately designed to ensure that direct success must come from actions with corrected coordinates from visual re-aiming or bounded step-local grounding repair.  All main experiments share the same framework including ActionLens action schema, OSWorld initialization environment, and artifact-verification mechanism. We use models and coding agent clients including Codex, Claude Code, and Kimi Code to conduct the main experiment. In addition, a baseline method Synapse~\citep{zheng2024synapse} which converts action-trajectory to augmented action planner was reported for naive comparison between planning enhancement and our proposed replayable execution methods.

The primary reported outcomes follow from these two-pass paired experiment. For the first pass, we report the token consumption, running time, as well as memory consolidation consumption as a reference for GUI agent executing the benchmark tasks. For the second pass, we reported correct memory recall rate, replay success rate, target-reaiming hit rate to demonstrate EchoPath's task validity. For efficiency, we report second-pass token consumption and task execution time to directly evaluate the efficiency advantage of replayable memory for industrial-scale tasks. The computation configurations for our main experiments are reported in Supplementary Materials Section C. 

\subsection{Core Functionality Analysis}
Beyond the main experiments, we conducted controlled analyses to test whether five crucial functional modules behave as expected, including memory selection, visual rebinding, storage, state compatibility, and flexible-input handling.

\paragraph{Retrieval stress test.}
We tested memory selection with an offline retrieval stress test without action execution. The source library is the current anchored consolidated OSWorld active memory entries with 0, 50, 100, 200, and 500 added non-target active trajectory records. Each added distractor is cloned from a real non-evaluation task and retains its own task intent and query phrases, so the test adds realistic library clutter rather than obviously irrelevant labels. Retrieval is performed through EchoPath's task retrieval module, using deterministic task-intent probes derived from the stored intent after removing task-id scaffolding and paraphrasing common action phrases, evaluating both retrieval accuracy and latency.

\paragraph{Image-based Target-Reaiming (IBTR) Algorithms.}
From the validated first-pass consolidated memory repository, we sampled 200 coordinate-based action cases with stored target crops and the corresponding pre-action screenshots. For each case, the target crop is matched back into the pre-action screenshot and the predicted click point is compared with the documented ActionLens coordinate. We ran two conditions. The first uses the original 1920$\times$1080 screenshot to test self-consistency of the stored visual evidence. The second is an anti-coordinate-copy condition: it applies a random resolution transform to each screenshot by resizing it with a uniform scale from 0.6 to 1.6, then rescales the documented coordinate by the same factor. This diagnostic measures coordinate binding precision and rejection behavior.

\paragraph{Local storage footprint.}
We evaluated the local storage footprint of each active consolidated memory by summing the path manifest, referenced action-node records, and path-specific visual evidence blobs. The analysis relates retained action steps to storage in megabytes and identifies whether large memories are driven by longer procedures or by visual evidence retained for audit and re-aiming. This measurement is performed on the consolidated repository and is intended to characterize how storage requirements may scale as enterprise deployments accumulate large libraries of reusable task procedures, without affecting memory selection or replay behavior.

\paragraph{State-gate calibration.}
We tested the replay start-state gate with 50 sampled memories with pointer-action visual evidence and required state contracts. Each memory is evaluated under three offline screenshot conditions. A compatible start uses the stored first-pass before-screenshot for the first pointer action. A partially changed start uses the before-screenshot for the first required pointer after optional startup or dependency-guard steps, which tests whether the gate can accept a task state that has already passed replay prerequisites. An incompatible start uses a first-pass screenshot from a different application whenever possible. The state gate receives the same memory contracts and visual-binding records used by replay. We reported the accuracy of the replay start-state gate among these three conditions to ensure the robustness of designed gating mechanism. 

\paragraph{Flexible-input rebinding.}
We tested replay with flexible input adaptation such as different text inputs or changed numbers on 50 active memories with declared \texttt{flexible\_action\_inputs}. For each memory, the replay task is materialized offline and paired with two synthetic reasoner responses. The valid condition modifies only the declared \texttt{action\_index} and \texttt{parameter\_path} fields, using replacement values matched to the recorded input kind, such as URL, filename, form value, or input text. The invalid condition attempts to change a fixed, non-flexible action field. The measured outcomes are whether all allowed replacements are applied, whether fixed replay actions remain unchanged, and whether non-flexible mutations are rejected.

\section{Results}
\begin{figure}[htbp!]
\centering
\includegraphics[width=0.98\linewidth]{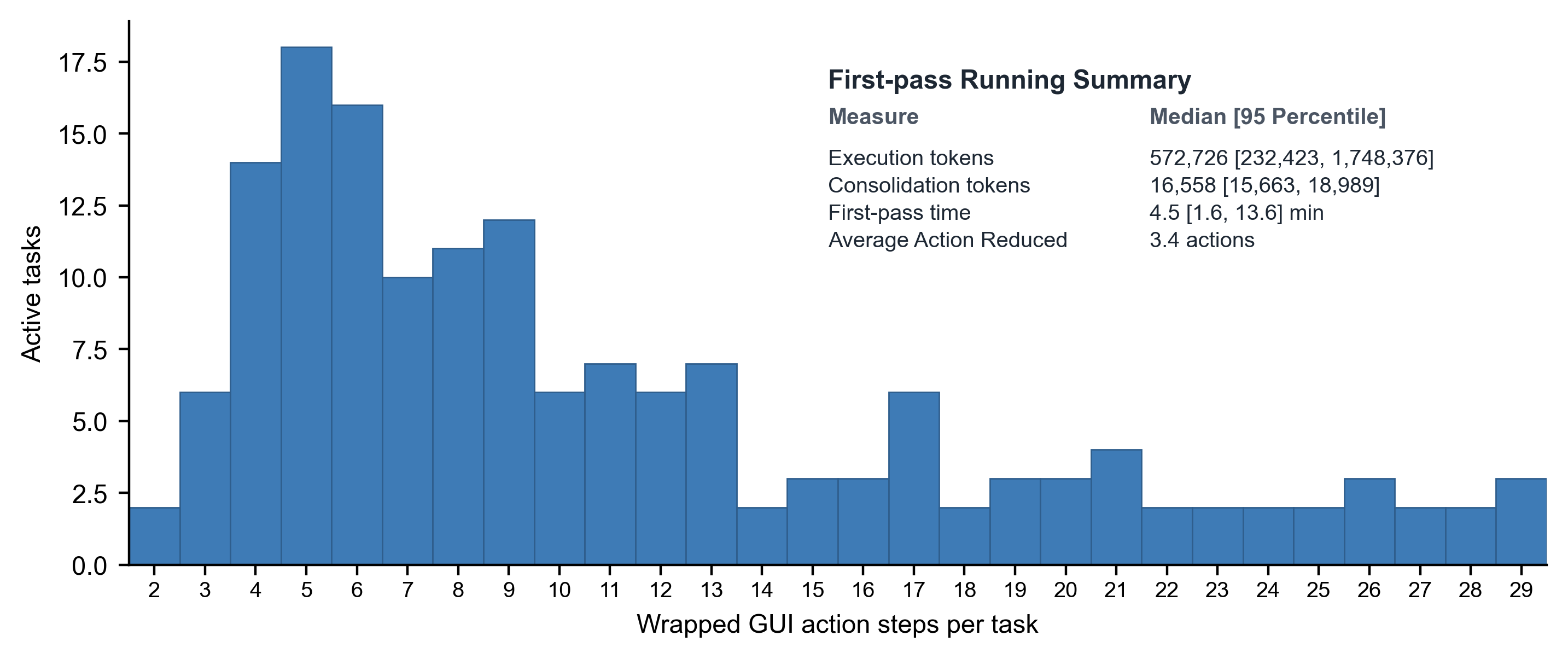}
\caption{First-pass memory construction for 159 active OSWorld memories. The histogram shows wrapped GUI action steps per task; the inset reports construction-cost medians and 0.025-0.975 quantiles.}
\label{fig:first-pass-overview}
\end{figure}

\begin{figure}[ht]
\centering
\begin{minipage}[t]{0.9\linewidth}
\centering
\textbf{(a)}
\includegraphics[width=\linewidth]{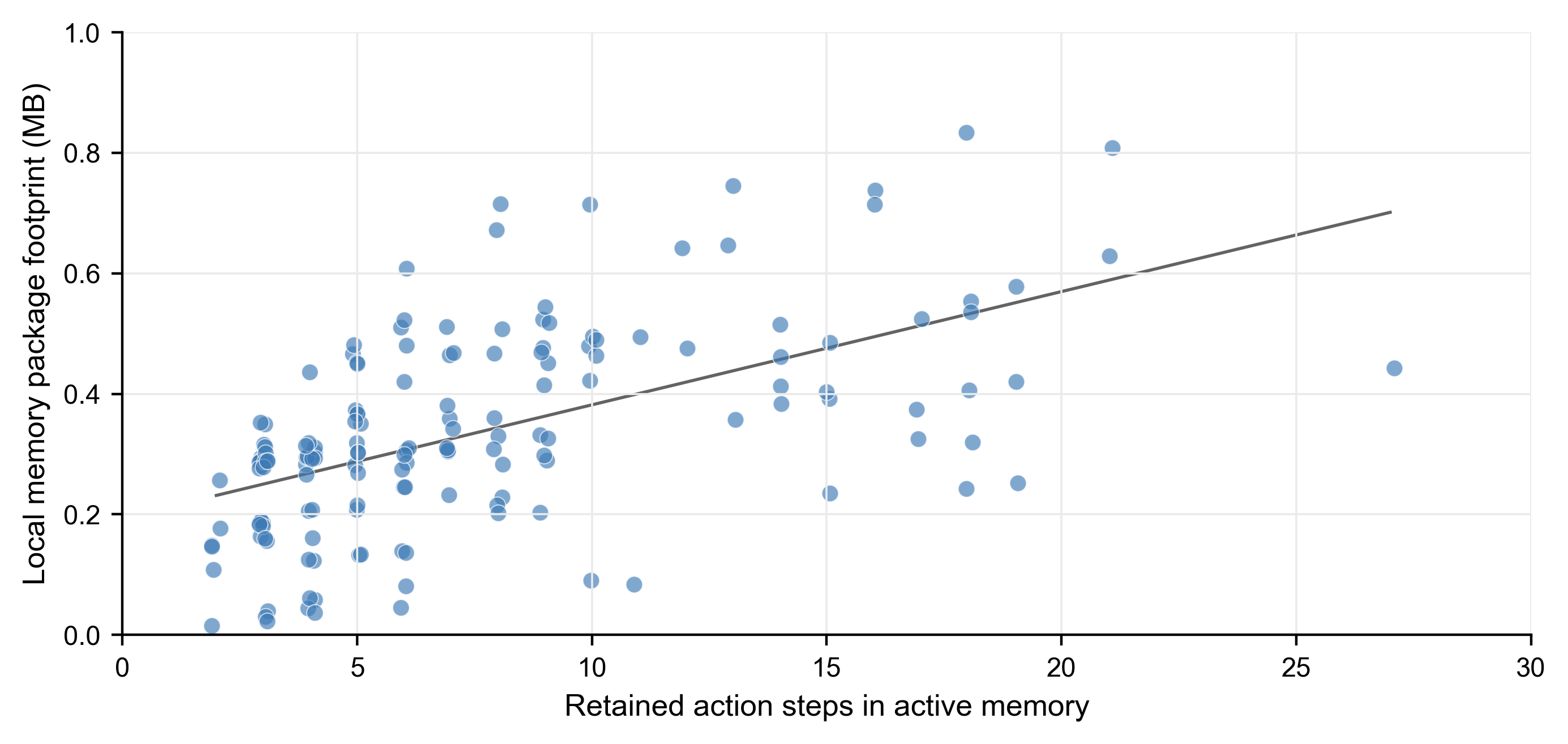}
\end{minipage}
\hfill

\begin{minipage}[t]{0.9\linewidth}
\centering
\textbf{(b)}
\includegraphics[width=\linewidth]{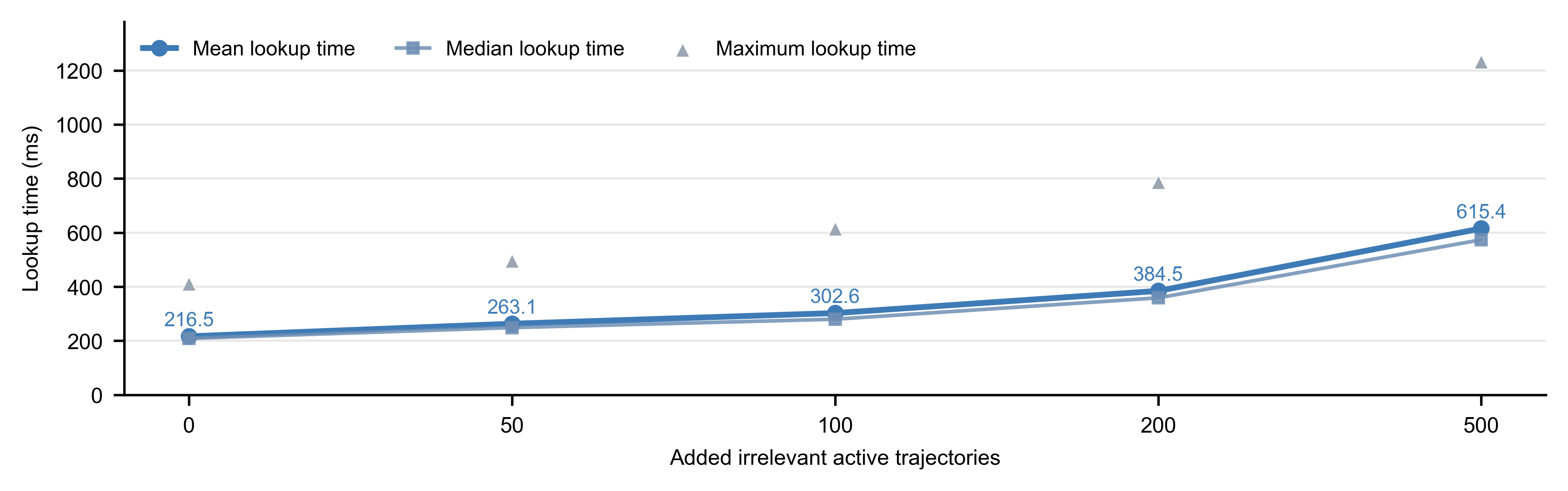}
\end{minipage}
\caption{Storage and retrieval scalability diagnostics. (a) Local memory footprint versus retained action steps. (coefficient = 0.019, intercept = 0.194) (b) Offline retrieval stress test reporting lookup time as the library size grows.}
\label{fig:memory-retrieval-panels}
\label{fig:memory-size-vs-steps}
\label{fig:retrieval-stress}
\end{figure}

The main analysis pool contained 159 active executable memories. Figure~\ref{fig:first-pass-overview} summarizes action counts and first-pass construction cost. Active tasks contained 2--29 recorded GUI actions with most tasks require 4 to 13 steps. The median task token cost is about 572k tokens, and the median execution time is about 4.5 minutes. Consolidation reduced about 30\% of the exploration steps to final memory trajectories. Figure~\ref{fig:memory-retrieval-panels}a reports the relation between number of retained steps in memory trajectories and local memory storage footprint, showing that the memory size remains small and manageable (less than 1 MB) even for long trajectory tasks.

\begin{table}[ht]
\centering
\caption{Second-pass memory replay results.}
\label{tab:second-pass-results}
\footnotesize
\setlength{\tabcolsep}{2pt}
\begin{tabular}{L{0.3\linewidth}C{0.12\linewidth}C{0.3\linewidth}C{0.3\linewidth}}
\toprule
Model & Success Rate & Execution Tokens Consumption & Second-pass time (seconds) \\
\midrule
codex-gpt-5.5-medium & 91.2\% & 20,370 [674--245,736]  & 127.5 [14.5--529.3] \\
claude-sonnet-5-medium  & 92.8\% & 30,503 [481--111,710] & 139.3 [19.5, 570.7] \\
kimi-K3-medium & 87.3\% & 19,928 [589--228,419] & 130.5 [15.2--627.6] \\
Synapse (codex-gpt-5.5-medium)  & 91.8\% & 586,386 [164,997--1,625,940] & 315.7 [110.1--1,272.5] \\
\bottomrule
\end{tabular}
\vspace{2pt}
\begin{minipage}{0.98\linewidth}
\end{minipage}
\end{table}

In matched second-pass evaluation, Codex EchoPath recalled the expected memory for all 159 tasks. It completed 145/159 matched tasks (91.2\%) and the median cost was 20,370 tokens and 127.5 seconds. Comparable results were also achieved by Claude Code and Kimi Code. As a planning augmentation baseline, Synapse reached comparable success rate at 91.8\% but used 586,386 median tokens and 315.7 seconds.

\paragraph{Retrieval stress test} Task-intent retrieval remained accurate under repository clutter (Figure~\ref{fig:memory-retrieval-panels}b). Correct recall stayed at 100\% accuracy with no wrong selections or misses as the active repository grew from 159 to 659 memories. Mean lookup time increased from 216.5 to 615.4 ms, median lookup time from 208.9 to 573.8 ms, and maximum lookup time to 1,232.0 ms. This test evaluates EchoPath's capability to precisely match memory with task with very low latency with scaled library size.

\paragraph{Target-reaiming test} The offline visual-binding diagnostic tested whether stored target crops could recover documented ActionLens pointer coordinates without GUI execution (Table~\ref{tab:visual-reaiming-diagnostic}). Original-screen matching accepted 95.5\% of 200 cases, and every accepted match had 0-pixel deviation. Under random scaling, the matcher accepted 190 cases with 188 within 2 pixels, 189 within 5 pixels, and 190 within 10 pixels away from ground truth. The accepted-match median deviation was 0 pixels, with a 2.5--97.5 percentile interval of 0--1.41 pixels. Rejections were concentrated in repeated or visually near-duplicate targets and were triggered mainly by the ambiguity margin.

\begin{table}[ht]
\centering
\caption{Offline visual re-aiming diagnostic on 200 coordinate-based actions.}
\label{tab:visual-reaiming-diagnostic}
\scriptsize
\setlength{\tabcolsep}{1.5pt}
\begin{tabular}{L{0.20\linewidth}L{0.12\linewidth}L{0.12\linewidth}L{0.12\linewidth}L{0.12\linewidth}L{0.18\linewidth}}
\toprule
Condition & Accept & $\leq$2 px & $\leq$5 px & $\leq$10 px & Deviation\\
\midrule
Original screen & 95.5\% & 95.5\% & 95.5\% & 95.5\% & 0.00 [0.00, 0.00]\\
Random scale & 95.0\% & 94.0\% & 94.5\% & 95.0\% & 0.00 [0.00, 1.41]\\
\bottomrule
\end{tabular}
\vspace{2pt}
\end{table}

\paragraph{State-gate and flexible-input diagnostics} 
The state gate accepted 49 compatible starts and 46 partially changed starts, and rejected 39 incompatible starts out of 50 testing cases each (Table~\ref{tab:state-rebinding-diagnostic}). False rejections came from low-confidence or ambiguous repetitive targets showing that a single visual probe may not be sufficient for generic controls. Flexible rebinding accepted all 50 valid declared text substitutions and rejected all 50 non-flexible mutations with 100\% accuracy.

\begin{table}[htbp!]
\centering
\caption{Offline state-gate and flexible-input diagnostics.}
\label{tab:state-rebinding-diagnostic}
\scriptsize
\setlength{\tabcolsep}{2pt}
\begin{tabular}{L{0.26\linewidth}L{0.26\linewidth}C{0.16\linewidth}C{0.12\linewidth}}
\toprule
Component & Condition & Expected Action & Accuracy\\
\midrule
State Gate & Compatible start & Accept & 98\%\\
State Gate & Partially changed start & Accept & 92\%\\
State Gate & Incompatible start & Reject & 78\%\\
\midrule
Flexible Rebinding & Valid flexible variant & Accept & 100\%\\
Flexible Rebinding & Invalid non-flex mutation & Reject & 100\%\\
\bottomrule
\end{tabular}
\vspace{2pt}
\end{table}

\FloatBarrier

\section{Discussion and conclusion}

EchoPath provides a concrete path from successful first-run GUI work to reusable second-run execution. Its core contribution is representing recurrent GUI tasks as validated, inspectable, and replayable action memory with explicit gates for reuse. The framework provides the IBTR algorithm as the core bridge to connect agent runtime history to execution-level replayable memory that bypasses grounding steps as well as a model-agnostic harness that connects retrieval and compatibility gating for memory management. This makes memory a controllable systems layer for reducing repeated model reasoning and grounding while preserving fallback when direct replay is unsafe.

The present evidence supports a system claim with EchoPath: when a task recurs under compatible state conditions, validated execution can replace repeated observe--plan--ground--act loops rather than merely provide extra context for them. The main practical implication is computational efficiency across models and agentic frameworks. EchoPath sits outside the host model and wraps GUI execution through a common action boundary, so the same replayable memory layer can in principle serve different coding agents, desktop agents, or model back ends without retraining the memory itself. This design is therefore especially relevant for employee work computers, where organizations may prefer local execution records, minimal hardware requirements, and reduced transmission of private screen or task context to online models. EchoPath also suggests a different scaling path for enterprise automation memory. A replayable memory is not only a cache entry for one agent session, but a versioned operational artifact with task intent, application context, state assumptions, action evidence, validation provenance, and lifecycle state. As repositories grow, such memories could become company-level digital assets: reusable procedures for recurring forms, reports, records, exports, and application configurations. 

EchoPath is best suited to stable work environments with controlled application versions, window systems, browser profiles, fonts, themes, and software configurations since replay-time visual re-aiming remains vulnerable to toolbar rearrangements, localization, responsive layouts, display scaling, and near-duplicate interface elements. The present diagnostics establish retrieval, offline visual binding, start-state gating, flexible-input enforcement, and storage behavior, but they do not yet establish live robustness under broad interface drift or enterprise-scale deployment. Another limitation is that memory acquisition currently depends on a first-pass agent attempting the task from the given instruction, rather than an interactive demonstration or user-guided recording procedure. As a result, gathering a replayable memory can still involve exploratory actions, failed attempts, and full task completion before consolidation, which may be inefficient for complex long-trajectory tasks.

Future work will focus on improving the EchoPath framework on the replay layer to incorporate stronger semantic UI understanding and richer state contracts so that memories remain usable under layout changes, software-version drift, localization, and display-scaling variation. The memory-acquisition phase could move beyond autonomous first-pass agent exploration toward interactive demonstrations, user-guided recording, and incremental consolidation, which would reduce the cost of collecting long-trajectory procedures. The repository layer could also support more robust branch/merge repair policies, privacy-preserving storage, and long-term governance for organizational-scale memory libraries.

\newpage
\bibliographystyle{unsrtnat}
\bibliography{references}

\clearpage
\appendix
\setcounter{table}{0}
\setcounter{figure}{0}
\renewcommand{\thesection}{\Alph{section}}
\renewcommand{\thesubsection}{\thesection.\arabic{subsection}}
\renewcommand{\thetable}{S\arabic{table}}
\renewcommand{\thefigure}{S\arabic{figure}}
\renewcommand{\theHsection}{\Alph{section}}
\renewcommand{\theHsubsection}{\Alph{section}.\arabic{subsection}}
\renewcommand{\theHtable}{S\arabic{table}}
\renewcommand{\theHfigure}{S\arabic{figure}}

\begin{center}
{\LARGE\bfseries Supplementary Materials for "EchoPath: Execution-Level Replayable Memory for GUI Agents"\par}
\end{center}

\section{Related Work}

\subsection{Computer-use agents and benchmarks}

OSWorld evaluates multimodal agents on open-ended tasks in real operating-system environments and uses execution-based task checks \citep{xie2024osworld}. Windows Agent Arena extends this style of evaluation to a scalable Windows environment \citep{bonatti2025windowsarena}, while AndroidWorld supplies parameterized mobile tasks with programmatic initialization and success checks \citep{rawles2025androidworld}. OSWorld 2.0 shifts the emphasis toward long-horizon, cross-application workflows and finer-grained progress evaluation \citep{yuan2026osworld2}. OSWorld-Human complements outcome accuracy with human-reference trajectories and efficiency metrics, and identifies online model calls as a major source of computer-use latency \citep{abhyankar2025osworldhuman}. Together, these benchmarks motivate EchoPath's use of reproducible initial states, external outcome checks, and a cost ledger for second-run execution.

Cradle establishes a general computer-control setting based on screenshots and keyboard or mouse actions, with explicit planning, skill, reflection, and memory modules \citep{tan2024cradle}. Agent S combines an agent-computer interface with experience-augmented hierarchical planning and retrieves external knowledge and internal experience \citep{agashe2024agents}. EchoPath adopts the same broad computer-use setting, but isolates recurrence as the experimental object: whether a previously successful procedure can be stored, rebound, replayed, and judged by its final artifact under matched first-run and second-run conditions.

A further per-step cost in these loops is GUI grounding, the binding of a planned action to concrete screen coordinates or interface elements. This step sustains its own line of work: Set-of-Mark overlays numbered visual markers so a model can refer to targets by index \citep{yang2023setofmark}, OmniParser parses a screenshot into localizable elements for pure-vision agents \citep{lu2024omniparser}, and dedicated grounding models localize elements directly from pixels-SeeClick \citep{cheng2024seeclick}, CogAgent \citep{hong2024cogagent}, and OS-Atlas \citep{wu2024osatlas}. EchoPath instead treats this recurring grounding cost as a stored episode: replay is designed to re-aim stored target evidence to the current screenshot through image matching, with an agentic fallback when binding fails.

\subsection{Experience-augmented planning and procedural memory}

ReAct established an interleaved reasoning-and-action loop for language agents \citep{yao2023react}. Non-parametric experience learning retains information across such loops as context for later decisions. Reflexion stores verbal feedback in an episodic buffer \citep{shinn2023reflexion}, and ExpeL extracts natural-language insights and recalls prior experiences at inference time \citep{zhao2024expel}. RAP retrieves contextually related multimodal experience for planning \citep{kagaya2024rap}; Agent Workflow Memory induces recurring textual workflows \citep{wang2024awm}; and Memp distils trajectories into fine-grained instructions and script-like abstractions \citep{fang2025memp}. LEGOMem similarly decomposes workflow trajectories into role-specific procedural memories for orchestrators and task agents \citep{han2026legomem}. The AFTER benchmark broadens this line by evaluating procedural-skill transfer across enterprise tasks, roles, and model backbones \citep{belikova2026procedural}. A separate, more architectural line treats memory as a general substrate for language agents: MemGPT pages a virtual context across hierarchical storage tiers \citep{packer2023memgpt}, Generative Agents organizes an episodic stream over which reflection and retrieval operate \citep{park2023generative}, MemoryBank couples storage with Ebbinghaus-inspired forgetting curves \citep{zhong2024memorybank}, and A-MEM maintains dynamically linked, evolving agentic memories \citep{xu2025amem}.

Other agents store callable abstractions rather than prose alone. Voyager maintains an expanding library of executable code skills for embodied tasks \citep{wang2024voyager}, while SkillWeaver discovers, practises, and distils website interactions into reusable APIs \citep{zheng2025skillweaver}. These systems clarify an important boundary: executable skills can still be selected or composed inside fresh model-driven planning. EchoPath instead tests whether a complete, artifact-validated action episode can bypass fresh per-step planning and grounding when compatibility and GUI re-aiming checks pass.

\subsection{Programming by demonstration and replay}

Execution-level reuse predates language-model agents. Sikuli locates GUI targets from screenshot patterns and drives mouse and keyboard actions through a visual scripting interface \citep{yeh2009sikuli}. Ringer turns user demonstrations into replay scripts and improves robustness to webpage changes by relying on relatively stable user-facing interfaces \citep{barman2016ringer}. WebRobot formalizes web robotic process automation as program synthesis from action demonstrations and couples speculative synthesis with validation \citep{dong2022webrobot}. These systems establish the classical record-and-replay lineage behind EchoPath's executor and re-aiming layer. EchoPath adds agent-generated trajectories, retrieval from an accumulated memory pool, external artifact gates, and an explicit lifecycle for promotion, repair, branching, merge, and retirement.

\subsection{Memory for GUI operation}

GUI memory has rapidly diversified. EchoTrail-GUI builds a reward-critic-filtered bank of successful trajectories and injects retrieved trajectories as in-context guidance \citep{li2026echotrail}. Continuous Memory encodes visual trajectories as fixed-length latent representations \citep{wu2025continuous}; HyMEM couples symbolic graph nodes with continuous trajectory embeddings and multi-hop retrieval \citep{zhu2026hymem}; and MementoGUI learns online selection, compression, writing, and retrieval over textual and region-level visual evidence \citep{zeng2026mementogui}. FocusMem separates latent-memory content, state-conditioned readout, and a trust gate \citep{zhang2026focusmem}. UI-Mem stores workflows, subtask skills, and failure patterns as parameterized templates for mobile-agent reinforcement learning \citep{xiao2026uimem}. Active Task Driving Memory instead maintains task-relevant values as an evolving execution state \citep{liu2026atmem}. Darwinian Memory decomposes GUI trajectories into reusable units and uses observed utility to retain useful paths and suppress risky ones without memory-model training \citep{mi2026darwinian}.

Executable Agentic Memory is the closest conceptual neighbor: it represents GUI routines in a structured knowledge graph and plans retrieval and execution with value-guided search \citep{qin2026eam}. EchoPath does not claim a more general memory algorithm. Its distinction is a lightweight systems and evaluation point: transparent execution-level action programs, replay-time GUI re-aiming, pre-planning replay, current-workspace rebinding, and artifact-evaluator evidence, without memory-model training.

\subsection{Verification, visual evidence, and runtime state}

Persistent memory can amplify mistakes if failed or misleading experiences are consolidated. VerificAgent treats memory as an alignment surface and adds expert seeding and post-hoc human fact checking before deployment \citep{nguyen2025verificagent}. Action-Grounded Visual Memory (AGMem) shows that full-screen visual history can worsen some GUI-agent failure modes and instead stores local image crops associated with successful actions or recovery \citep{choi2026agmem}.

OpenRath addresses a different systems layer. Its first-class \term{Session} carries conversation chunks, workspace placement, lineage, token usage, tool evidence, and memory-event boundaries so that branch, merge, inspection, and replay are explicit runtime operations \citep{wen2026openrath}. The OpenRath report deliberately leaves memory quality to follow-on evaluation. It is consequently a complementary runtime-state comparison, not a GUI-memory performance baseline: OpenRath broadens auditable runtime state, whereas EchoPath operationalizes and evaluates one narrow executable-memory path.

\section{Methodology Details}
\subsection{ActionLens: wrapped GUI operation}

ActionLens is the execution boundary that makes GUI operation observable and replayable. The module is needed because host-agent logs, private reasoning traces, and SDK event streams are not stable enough to serve as the source of replay memory. A host agent may still plan the task and choose actions, but ActionLens is responsible for recording the normalized GUI operations that actually touch the desktop.

ActionLens is implemented as a controller-compatible wrapper around the existing desktop back ends. It exposes the action space $\mathcal{A}$ and preserves the method surface of the underlying controller, so it can wrap either a local PyAutoGUI controller or a remote VM controller. Each action object $a_t=(\alpha_t,\theta_t)$ is executed through the underlying controller, while ActionLens records the evidence object $z_t$. This design also gives any agentic framework the same function-call interface without making replay depend on their internal trace formats.

Each wrapped run creates an immutable evidence directory containing a manifest, action-step JSONL, standalone-observation JSONL, lifecycle-event JSONL, and screenshot assets. The manifest records the run identifier, source agent, task goal, controller name, status, output paths, and record counts. Standalone observations are stored separately because a host agent may inspect the GUI before deciding whether an action is needed.

For each action step, ActionLens follows a fixed capture protocol. It allocates a monotonic step identifier, captures a screenshot before observation, records the requested action and source metadata, executes the normalized primitive, waits for a short settling interval, and captures another screenshot after observation. The resulting $z_t$ record stores UTC timestamps, measured duration, action parameters, executor output, error text if execution failed, before and after observations, and coordinate context. The coordinate context includes logical screen size, mouse position, screenshot image size, optional screenshot region, and inferred pixel-to-controller scale factors. These fields are later used by action replay to avoid treating raw coordinates as portable facts.

Failure handling is part of the record format. If an action raises an exception, ActionLens still attempts to capture the after state, writes a step with status \texttt{error}, and returns the error to the caller. Failed first-run traces can therefore support diagnosis and repair, but they are not promoted to active replay memory by default. Promotion remains gated by external artifact evaluation rather than by successful GUI primitive execution alone.

\subsection{Memory retrieval, and consolidation}

\paragraph{Task retrieval.}
Retrieval is a gated selection problem over $\mathcal{M}_{\mathrm{active}}$. Each active memory is represented for retrieval by its stored task intent, query phrases, application label, reasoning summary, preconditions, and expected effects. The repository converts both the current request and each memory record into normalized content terms: text is case-folded and whitespace-normalized, common function words are removed, and frequent GUI-task verb variants such as \emph{save}/\emph{store}, \emph{find}/\emph{search}, and \emph{replace}/\emph{overwrite} are canonicalized. For a new task $\tau'=(q',\xi',V')$, the retriever builds an intent-query set $I(\tau')$ from the user goal, optional caller-supplied intent phrases, nested desktop-task goal and summary fields, and instruction-like payload fields. EchoPath computes an pairwise intent score $s_I(m,\tau')\in[0,1]$ between query and memories, from query-term coverage, overlap coefficient, token-level F1, and ordered bigram overlap. The candidate set is
\[
\mathcal{C}(\tau')=
\{m\in\mathcal{M}_{\mathrm{active}}: s_I(m,\tau')\geq\lambda\},
\]
where the current first-stage threshold is $\lambda=0.18$. Lookup proceeds in two stages. First, the retriever ranks active paths by $s_I$ for every phrase in $I(\tau')$, keeps the strongest match for each \texttt{path\_id}, and de-duplicates repeated paths found through multiple phrases. Second, it loads each surviving path record and its action-node records for compatibility evaluation. This second stage is what turns an intent hit into a replayable memory hit: a path with similar task intent can still be rejected because it is no longer active, names an incompatible application, lacks positive artifact validation, has a non-viable reasoning report, contains unsupported action primitives, lacks required visual evidence for replay, or falls below the replay score threshold. The application gate is permissive when either side lacks an application label, but requires an exact normalized match when both the current task and the memory name one. The lookup summary records searched intent phrases, candidate count, rejected paths, failed gate names, selected path, matched phrase, intent score, and compatibility score, so retrieval decisions can be audited after a run.

Each candidate is scored only after explicit gates are evaluated. We use binary gates for active lifecycle status $g_{\mathrm{life}}$, sufficient task-intent match $g_{\mathrm{intent}}=\mathbf{1}[s_I\geq0.32]$, desktop task type $g_{\mathrm{type}}$, application agreement $g_{\mathrm{app}}$, artifact validation $g_{\mathrm{val}}$, reasoning viability $g_{\mathrm{reason}}$, and supported action primitives $g_{\mathrm{act}}$. The compatibility score is
\[
\begin{aligned}
S(m,\tau',x_0) ={}& 0.15g_{\mathrm{life}}
 + 0.40s_I(m,\tau')
 + 0.05g_{\mathrm{type}}
 + 0.10g_{\mathrm{app}}
 + 0.15g_{\mathrm{val}} \\
&+ 0.10g_{\mathrm{reason}}
 + 0.05g_{\mathrm{act}} .
\end{aligned}
\]
A memory is replayable only if all gates pass and $S(m,\tau',x_0)\geq\theta$, where the current threshold is $\theta=0.75$. The selected memory is
\[
m^\star=\arg\max_{m\in\mathcal{C}(\tau')} S(m,\tau',x_0)
\]
If no memory satisfies the gates, EchoPath records a miss or rejection and starts regular planning with ActionLens operation.

\paragraph{Consolidation.}
Consolidation converts an evaluated trace into a candidate episode and decides whether it should become active memory. The lifecycle follows Git-style operations. A new evaluated trace begins as a candidate. A validated and viable candidate can be promoted to active memory. If later reuse fails under changed state, the episode can be quarantined or forked into a repaired branch. Equivalent prefixes or subpaths can be merged after evaluation, and stale or unsafe memories can be deprecated without deleting their audit trail. The repository also supports deterministic diff and export operations: diff reports shared prefixes and divergent branches, while export materializes an ordered task for action replay. 

For an recorded action trajectory, the repository first normalizes each supported GUI and maps it to an action node. Each node is identified by a deterministic signature over action type, normalized parameters, and GUI-component signature. Tasks that begin with the same GUI operations therefore share prefix nodes, while task-specific downstream edits form branches. The resulting structure consists an action graph with explicit child links, source-step lineage, reuse counts, and per-path lifecycle references. Model-assisted consolidation can also compact a successful wrapper trace before ingestion. The consolidator selects a non-empty ordered subsequence of successful steps, drops exploratory or redundant actions with recorded reasons, and returns the reasoning fields used by the promotion gate.

Visual evidence is stored separately from path metadata. Full before/after screenshots and derived GUI target crops are written as content-addressed blobs using cryptographic hashes. Episode and node records contain only references, dimensions, coordinate frames, crop boxes, source run identifiers, and observation metadata. This keeps the audit trail lossless while avoiding duplicate screenshot storage when multiple memories reuse the same GUI evidence.

Promotion is gated by artifact evidence and consolidation reasoning. The external gate must report that the produced artifact passed and must name an evaluator. The reasoning gate records task intent, query keys, required preconditions, expected effects, blockers, and a pass/fail promotion recommendation. A candidate is rejected if any action failed, if an action is unsupported, if artifact validation is absent or negative, if the reasoning report does not mark the path viable for promotion, or if blockers remain. Rejected candidates stay in the repository for diagnosis or repair, but only active memories are exposed to default retrieval.

\subsection{Image-based target-reaiming Algorithm}
\label{sec:target-reaiming}

The image-match target-reaiming (IMTR) algorithm is the deterministic local binding procedure used when a stored pointer action must be aimed at the current GUI. IMTR is narrower than general GUI grounding. It does not infer a target from language; it asks whether the visual target used by a previously validated action appears again, clearly enough to reuse the action. For action $a_t$, the memory repository stores one or more component crops $c_{t,j}$ centered on the source action point, the click ratio $r_{t,j}=(r_x,r_y)$ inside each crop, the original controller coordinate $\theta_t^{xy}$, the source observation metadata, and pixel-to-controller scale metadata. The current implementation considers \texttt{target}, \texttt{context}, and \texttt{wide\_context} crop variants, with the target crop tried first.

The intuition is to treat each stored crop as a visual fingerprint. A crop around a button, tab, menu item, or icon contains a spatial pattern of edges, text strokes, background contrast, and local layout. IMTR slides this fingerprint over the current full-screen observation image $I'_t$ and tests whether any location has the same pattern. Because the current screen can be resized, IMTR also tests each crop over a scale set
\[
\mathcal{S}=\{s_{\min},s_{\min}+\Delta_s,\ldots,s_{\max}\},
\]
where each $s$ resizes a crop before matching. For a crop with width $w$ and height $h$, IMTR uses scaled dimensions $\operatorname{round}(sw)$ and $\operatorname{round}(sh)$. The current implementation resizes the stored crop with area interpolation when $s<1$ and bilinear interpolation when $s\geq1$. The current full-screen screenshot is not globally resized; each resized crop is compared against same-sized patches in the current screenshot.

We write one candidate match as $d=(j,s,p)$, where $j$ selects the stored crop, $s$ is the tested scale, and $p$ is a top-left position in the current full-screen screenshot. Let $\mathcal{D}_t$ be the set of all such candidates whose resized crop fits within $I'_t$. For candidate $d$, let $C_d$ be the vectorized resized crop in the matching representation, grayscale by default, and let $Q_d$ be the vectorized current-screen patch under the same position and size. IMTR chooses the scoring rule from the visual variation in the stored crop. Let $\sigma(C_d)$ be the maximum per-channel standard deviation of the preprocessed crop, which is a single standard deviation in grayscale mode. The current implementation uses a flatness threshold $\sigma_{\mathrm{flat}}=1.0$ pixel-value unit. When $\sigma(C_d)\geq\sigma_{\mathrm{flat}}$, the crop has enough internal contrast for zero-mean normalized correlation:
\[
\gamma(d)=
\max\!\left(0,
\frac{\langle Q_d-\bar{Q}_d,\, C_d-\bar{C}_d\rangle}
{\|Q_d-\bar{Q}_d\|_2\|C_d-\bar{C}_d\|_2}
\right),
\]
where negative correlations are clamped to zero before thresholding. This score is computed from aligned pixels, but it functions as a pattern-matching score because all pixels vote together. A correct target receives a high score when dark regions, bright regions, edges, and text strokes rise and fall in the same spatial positions as the stored crop. A wrong patch receives a lower score because its pixel-wise agreements are weaker or cancel across the crop.

When $\sigma(C_d)<\sigma_{\mathrm{flat}}$, the crop is treated as nearly constant and the correlation denominator is unstable. IMTR then uses normalized squared difference and converts it to the same similarity orientation:
\[
\gamma_{\mathrm{flat}}(d)
=
1-
\min\!\left(1,
\frac{\|Q_d-C_d\|_2^2}
{\|Q_d\|_2\|C_d\|_2+\epsilon}
\right).
\]
This explicit flatness test prevents plain buttons or uniform toolbar regions from producing undefined correlation scores. In both branches, larger values mean that the current patch is more similar to the stored crop.

The best visual match is
\[
d^\star=(j^\star,s^\star,p^\star)=
\arg\max_{d\in\mathcal{D}_t}\gamma(d),
\qquad
\gamma^\star=\gamma(d^\star).
\]
High similarity alone is not sufficient, because GUI screens often contain repeated controls, icons, tabs, or list entries. IMTR therefore asks a second question: is the best match clearly better than the next plausible match? To avoid counting small shifts of the same target as separate alternatives, IMTR defines a suppression neighborhood around the best candidate. For $d^\star=(j^\star,s^\star,p^\star)$, let
\[
N(d^\star)=
\left\{p:
|p_x-p^\star_x|\leq\left\lfloor\frac{w(c_{t,j^\star}^{(s^\star)})}{2}\right\rfloor,\,
|p_y-p^\star_y|\leq\left\lfloor\frac{h(c_{t,j^\star}^{(s^\star)})}{2}\right\rfloor
\right\},
\]
with this neighborhood clipped to the score-map boundary. Candidate $d=(j,s,p)$ is treated as overlapping the best match when its top-left position $p$ lies in $N(d^\star)$. IMTR then records the strongest remaining non-overlapping score:
\[
\gamma^{(2)}=
\max_{\substack{d\in\mathcal{D}_t:\\
p(d)\notin N(d^\star)}}
\gamma(d),
\]
where $p(d)$ is the top-left screen position of candidate $d$, and $\gamma^{(2)}=0$ if no second candidate exists. The ambiguity margin is
\[
\Delta\gamma=\gamma^\star-\gamma^{(2)}.
\]
The match is accepted only when both the absolute confidence and the margin pass:
\[
\gamma^\star\geq\eta
\quad\mathrm{and}\quad
\Delta\gamma\geq\eta_{\mathrm{margin}}.
\]
The current replay configuration uses $\eta=0.78$, $\eta_{\mathrm{margin}}=0.02$, $s_{\min}=0.5$, $s_{\max}=2.0$, and $\Delta_s=0.05$ unless overridden. If either condition fails, IMTR rejects the replay step before any pointer event is emitted, allowing step-local repair or higher-level fallback to take over.

When the match is accepted, the predicted target point in screenshot pixels is
\[
c^\star=c_{t,j^\star}^{(s^\star)},\qquad
u'_t=p^\star+\left(r_x\,w(c^\star),\,r_y\,h(c^\star)\right).
\]
If the full-screen observation has a different pixel scale from the controller coordinate frame, EchoPath converts $u'_t$ through the observation metadata and current screen size before emitting the action:
\[
\tilde{\theta}_t^{xy}
=
\left(
\operatorname{round}\!\left(\frac{u'_{t,x}}{\sigma_x}\right),
\operatorname{round}\!\left(\frac{u'_{t,y}}{\sigma_y}\right)
\right),
\]
where $\sigma_x,\sigma_y$ are the current image-to-controller scale factors inferred from the observation image size and logical screen size. Coordinates outside the current screen are rejected. The final replay action keeps the original primitive and non-coordinate parameters but replaces the pointer coordinates with $\tilde{\theta}_t^{xy}$. This formulation makes the stored coordinate useful as evidence for crop construction, while the replay-time coordinate is determined by visual evidence in the current full-screen observation.

\begin{center}
\begin{minipage}{0.98\linewidth}
\small
\hrule
\vspace{2pt}
\noindent\textbf{Algorithm 1 Image-match Target-Reaiming}
\vspace{2pt}
\hrule
\vspace{2pt}
\setlength{\tabcolsep}{0pt}
\renewcommand{\arraystretch}{1.02}
\begin{tabular}{@{}r@{\hspace{0.45em}}>{\raggedright\arraybackslash}p{0.88\linewidth}@{}}
1: & \textbf{Input:} stored pointer action $a_t$, crop candidates $\{c_{t,j},r_{t,j}\}$, current full-screen observation $(I'_t,o'_t)$, scale set $\mathcal{S}$, thresholds $\eta$, $\eta_{\mathrm{margin}}$, and $\sigma_{\mathrm{flat}}$\\
2: & $\textsc{Best}\leftarrow\emptyset$; $\textsc{Second}\leftarrow0$\\
3: & Prepare the current screenshot in the selected match mode, grayscale by default\\
4: & \textbf{for} each crop candidate $c_{t,j}$ \textbf{do}\\
5: & \hspace*{1.2em}\textbf{for} each scale $s\in\mathcal{S}$ \textbf{do}\\
6: & \hspace*{2.4em}$C_s\leftarrow$ resize $c_{t,j}$ to $\operatorname{round}(s w)\times\operatorname{round}(s h)$ using area interpolation if $s<1$ and bilinear interpolation otherwise\\
7: & \hspace*{2.4em}\textbf{if} $C_s$ is larger than the current screenshot \textbf{then} continue\\
8: & \hspace*{2.4em}For every valid full-screen position $p$, form candidate $d=(j,s,p)$; compute correlation score if $\sigma(C_d)\geq\sigma_{\mathrm{flat}}$, otherwise compute $\gamma_{\mathrm{flat}}(d)$\\
9: & \hspace*{2.4em}Update $\textsc{Best}$ and the strongest non-overlapping $\textsc{Second}$ candidate\\
10: & \hspace*{1.2em}\textbf{end for}\\
11: & \textbf{end for}\\
12: & Let $\textsc{Best}=(d^\star=(j^\star,s^\star,p^\star),\gamma^\star)$ and $\gamma^{(2)}\leftarrow\textsc{Second}$\\
13: & $\Delta\gamma\leftarrow\gamma^\star-\gamma^{(2)}$\\
14: & \textbf{if} $\gamma^\star<\eta$ or $\Delta\gamma<\eta_{\mathrm{margin}}$ \textbf{then}\\
15: & \hspace*{1.2em}Reject before issuing any pointer action\\
16: & \textbf{end else}\\
17: & $u'_t\leftarrow p^\star+(r_x\,w(c_{t,j^\star}^{(s^\star)}),\,r_y\,h(c_{t,j^\star}^{(s^\star)}))$\\
18: & $\tilde{\theta}_t^{xy}\leftarrow T_{\mathrm{img}\rightarrow\mathrm{ctrl}}(u'_t,o'_t,\xi')$\\
19: & \textbf{if} $\tilde{\theta}_t^{xy}$ is outside the current screen \textbf{then} reject\\
20: & \textbf{Output:} original action primitive with coordinates $\tilde{\theta}_t^{xy}$ and the score, margin, crop, scale, and transform report\\
\end{tabular}
\vspace{2pt}
\hrule
\end{minipage}
\end{center}

\section{Experiment Setup}
The experiments were conducted with Amazon AWS Virtual Machines. No GPU is required to initialize and run EchoPath with customized agentic frameworks.

\begin{table}[ht]
\centering
\caption{Primary OSWorld Worker Configuration}
\label{tab:osworld_worker}
\begin{tabular}{ll}
\toprule
\textbf{Component} & \textbf{Configuration} \\
\midrule
EC2 type & \texttt{t3.large} \\
CPU & 2 vCPUs \\
RAM & 8 GiB \\
Storage & 80 GiB \\
AMI & Ubuntu Server 24.04 LTS, amd64 \\
\bottomrule
\end{tabular}
\end{table}

\section{Prompts Used by Major Pipeline Modules}

This supplement records the fixed prompt text used by the major model-calling modules in the EchoPath pipeline. Runtime fields such as task identifiers, task instructions, screenshot paths, observation metadata, history summaries, and action lists are inserted into the JSON payloads shown by placeholder names.

\subsection{Desktop Next-Step Planning}

The desktop planner receives the current screenshot, recent action history, observation metadata, and task-specific guidance, then returns one GUI step.

\begin{PromptBox}
You are the EchoPath planning agent handling desktop execution planning. Review the screenshot and decide the single best next step to advance the task. Respond with JSON only using this shape: {"assessment": string, "completed": boolean, "next_step": object|null}. Use exactly one next_step at a time. Allowed next_step.action values are: click_target, double_click_target, move_target, click, move, type, paste_text, press, hotkey, scroll, wait. Every next_step must include an expected_outcome string describing what must be visibly true in the very next screenshot if the step succeeds. Use *_target actions when the exact pixel must be chosen from the image later by an image-grounding agent. For click_target, double_click_target, or move_target include a clear target_description and optionally target_text. For click or move without grounding, include integer x and y coordinates directly. For hotkey, include keys. For press, include key. For type or paste_text, include text. Prefer paste_text over type for exact text, multi-line text, structured data, JSON, formulas, or any full-file replacement, because it avoids partial keyboard entry mistakes. For scroll, include clicks. For wait, include seconds. Only return completed=true when the requested end state is directly visible in the current screenshot. Do not infer completion from Dock indicators, menu bar state, background app status, or assumed focus. For app, window, or browser launch tasks, completed=true requires the requested app or window to be visibly open and ready for interaction in the screenshot. If the task is unsafe to continue, return completed=true with next_step=null and explain why in assessment.

Task-specific guidance:
{planner_prompt}
\end{PromptBox}

When memory-enhanced fallback is enabled, the following text is appended to the same desktop planner instructions.

\begin{PromptBox}
Memory-enhanced replay fallback guidance:
You are not doing unconstrained full planning. A deterministic memory replay failed on the recorded action index shown in the payload, and your job is to pilot the same stored trajectory from the live screenshot. Use the trajectory as the primary plan. Prefer executing the current or next memory action, adapted to the live UI. You may perform one safe repair action to dismiss a transient blocker, re-open an earlier parent menu or dialog from the memory trajectory, or skip a stale optional prerequisite when a downstream memory target is already visible. Do not invent unrelated steps while a memory step or repair can advance the replay. When selecting a remembered visual target, include its memory\_target\_id or visual\_target\_id from the trajectory so the grounding layer can choose the matching reference crop. For remembered text, hotkey, press, scroll, or wait steps, preserve the stored parameters unless the live state makes a small adaptation clearly necessary. Mark completed=true only when the original user goal is visibly complete.

Memory reference crop images are attached after the live screenshot. Use them to identify remembered replay targets.
\end{PromptBox}

\subsection{Image Grounding}

The grounding agent receives the live screenshot, the planner's target description, and optional reference crops, then returns coordinates only when the target is clear.

\begin{PromptBox}
You are the EchoPath image grounding agent. Review the screenshot and locate the exact pixel coordinates for the requested target. Respond with JSON only using this shape: {"located": boolean, "x": integer|null, "y": integer|null, "confidence": number, "reason": string}. Only return coordinates when the target is clearly visible and you are confident. If the target is missing or ambiguous, return located=false with x=null and y=null.
\end{PromptBox}

When reference crops are attached for remembered replay targets, the following text is appended.

\begin{PromptBox}
Reference crop images are attached after the live screenshot. Use them as visual evidence for the exact memorized UI target.
\end{PromptBox}

When the grounding call is a bounded replay-step fallback, the following text is appended.

\begin{PromptBox}
This is a bounded replay-step fallback, not a full task plan. If the memorized target is visible, return its coordinates and set target_role="memory_target". If the target is hidden by a transient dialog, onboarding screen, keyring prompt, or similar blocker, you may instead return one safe dismiss/continue/cancel/close coordinate that should expose the same target, and set target_role="repair_action". Do not advance the broader task or invent multi-step plans.
\end{PromptBox}

\subsection{Step Verification}

The step verifier compares before and after screenshots for the executed GUI action.

\begin{PromptBox}
You are the EchoPath desktop step verification agent. Compare the before and after screenshots and determine whether the executed step achieved, advanced, or blocked its expected outcome. Respond with JSON only using this shape: {"assessment": string, "step_succeeded": boolean, "outcome_status": "succeeded"|"progressed"|"pending"|"blocked", "blocker": string|null}. Use outcome_status=succeeded and step_succeeded=true only when the action's expected outcome is directly visible in the after screenshot or proven by provided non-visual evidence. Use outcome_status=progressed when the exact expected outcome is not fully visible yet, but the UI visibly advanced toward the goal, such as a URL changing to a PDF route, a loading viewer appearing, a Save/Download dialog opening, or a provided download_evidence entry showing a new completed file. Use outcome_status=pending when the operation appears to be settling and there is no clear blocker yet. Use outcome_status=blocked when the screen is unchanged, the wrong window prevents the next safe step, a permission/error dialog blocks progress, or there is direct evidence the action failed. Do not infer success from Dock indicators, app running status, menu bar state, or assumed keyboard focus. For exact text, formula, or structured-data edits, do not mark the step succeeded merely because a requested substring is visible; check for obvious extra or missing characters, duplicate braces, syntax errors, stale text, or unsaved dirty indicators in the visible editor.

Task-specific verification guidance:
{verification_prompt}
\end{PromptBox}

\subsection{Goal Verification}

The goal verifier checks whether the requested desktop goal is complete in the current screenshot or by provided non-visual evidence.

\begin{PromptBox}
You are the EchoPath desktop goal verification agent. Review the screenshot and determine whether the user's requested desktop goal is visibly complete. Respond with JSON only using this shape: {"assessment": string, "goal_completed": boolean, "blocker": string|null}. Mark goal_completed=true only when the requested end state is directly visible and ready for interaction in the screenshot, or when provided non-visual evidence such as download_evidence proves that the requested downloaded file exists and is complete. Do not infer completion from Dock indicators, background app status, menu bar state, or assumed focus. For app, window, or browser launch tasks, the requested app or window must be visibly open in the screenshot. For exact text, formula, or structured-data edits, do not mark the goal complete merely because a requested substring is visible; reject obvious extra or missing characters, duplicate braces, syntax errors, stale text, or unsaved dirty indicators in the visible editor. If the goal is not visibly complete, set goal_completed=false and describe the blocker or missing evidence.

Task-specific verification guidance:
{verification_prompt}
\end{PromptBox}

\subsection{Memory Trajectory Consolidation}

After a successful first pass, the memory consolidator receives compacted action-step records and selects the replayable subsequence.

\begin{PromptBox}
Consolidate this successful GUI trajectory into a deterministic replay memory. Select only actions that are necessary and useful for reproducing the completed task. Preserve their original order. Drop exploratory clicks, corrections that were later undone or overwritten, redundant focus changes, no-op moves, unnecessary waits, and failed actions. Keep required navigation, focus acquisition, edits, save/submit actions, and waits only when they are necessary for a state transition. Preserve state-preparation dependencies even when they look unrelated to the final task, including clicks or keypresses that dismiss onboarding, sign-in prompts, tips, modal dialogs, popups, permission prompts, blockers, or other first-run UI state before the main task action. Treat a step as potentially dependent when the screen changes immediately after it, or when the screen changes between that step's after-screenshot and the next step's before-screenshot. Do not drop such a step unless a later retained action explicitly reaches the same UI state without relying on it. Never invent, rewrite, combine, or reorder actions. Return selected_step_ids as a non-empty subsequence of the supplied successful steps. Mark flexible_action_inputs for selected type or paste_text actions whose text/search/form value is task-specific and could be rebound for a closely related future task. Use parameter_path relative to the replay action, for example ["text"] for paste_text.text or type.text. Do not mark fixed navigation, save hotkeys, waits, coordinates, or structural actions as flexible unless the task explicitly makes that parameter the reusable variable. Explain every dropped step, and provide the promotion reasoning fields requested by the schema. Mark viable true only when the selected action subsequence is complete enough to replay the stated task under the listed preconditions. Set promotion_recommendation to exactly 'promote' or 'reject'; reject when the selected steps omit required task effects, contain unsupported/failed actions, or depend on unstated missing state. Put concrete failure reasons in blockers.

{"task_intent":"{task_intent}","application":"{application}","steps":[{compacted_successful_gui_steps}]}
\end{PromptBox}

\subsection{Replay-Time Flexible-Input Reasoning}

When a selected memory declares flexible text-like inputs, the replay reasoner decides whether the stored procedure can run as-is, should substitute only approved values, or must be rejected.

\begin{PromptBox}
Decide whether a stored GUI replay memory can execute the current task. The procedure is fixed except for the listed flexible_action_inputs. Return execute_as_is if the stored action values already satisfy the current task. Return modify_flexible_inputs if the same GUI procedure applies and only listed flexible values, such as search contents or input text, need replacement. Return reject if the current task requires changing any non-flexible action, different navigation, a different application, unsafe behavior, or information not available in the current task. Only produce modifications that target one of the provided flexible action_index + parameter_path pairs. Do not rewrite, add, remove, or reorder actions.

{"current_goal":"{goal}","current_task":{current_task_payload},"memory":{"path_id":"{path_id}","task_intent":"{memory_task_intent}","application":"{application}","matched_query_key":"{matched_query_key}","compatibility":{compatibility_report}},"actions":[{compact_replay_actions}],"flexible_action_inputs":[{flexible_action_inputs}]}
\end{PromptBox}

\subsection{Replay-Step Grounding Fallback}

When visual re-aiming fails but bounded step-local grounding fallback is enabled, the following guidance is appended to the grounding prompt for the current replay action.

\begin{PromptBox}
Replay is the execution priority. This fallback is only for the current memory replay action {action_index}; do not solve the whole task from scratch. Prefer the target from the reference crop. If a temporary blocker hides it, choose one safe dismiss/continue/cancel/close control and label it target_role='repair_action'. Otherwise label the target hit target_role='memory_target'.

Visual re-aiming failure: {reaiming_error}

Fallback attempt: {fallback_attempt}
\end{PromptBox}

\end{document}